\documentclass{article}
\PassOptionsToPackage{numbers, compress}{natbib}
 \usepackage[preprint]{neurips_2025}

\usepackage[utf8]{inputenc} 
\usepackage[T1]{fontenc}    
\usepackage{hyperref}       
\usepackage{url}            
\usepackage{booktabs}       
\usepackage{amsfonts}       
\usepackage{nicefrac}       
\usepackage{microtype}      
\usepackage{xcolor}         

\usepackage{multicol}
\usepackage[utf8]{inputenc}
\usepackage{graphicx}
\usepackage{colortbl}
\usepackage{pifont}
\usepackage{subcaption}

 \usepackage{amsmath}
\usepackage{booktabs}
\usepackage{multirow}

\definecolor{green}{RGB}{0,150,10}
\definecolor{blue}{RGB}{0,148,181}
\definecolor{orange}{RGB}{194,153,107}

\usepackage{xspace}
\newcommand{\OURS}{\textsc{EWMBench}\xspace} %
\newcommand{\Ours}{\OURS} %

\title{\OURS: Evaluating Scene, Motion, and Semantic Quality in Embodied World Models}

\author{%
Yue Hu$^{1,4,*}$ \\
\And
Siyuan Huang$^{2,*}$ \\
\And
Yue Liao$^{3}$ \\
\And
Shengcong Chen$^{1}$ \\
\And
Pengfei Zhou$^{1}$ \\
\And
Liliang Chen$^{1,\dag}$ \\
\And
Maoqing Yao$^{1,\diamond}$ \\
\And
Guanghui Ren$^{1,\diamond}$ \\
}

\begin{document}

\maketitle

\renewcommand{\thefootnote}{\fnsymbol{footnote}} 
\footnotetext[1]{\hspace{-1.5mm}$^{*}$Equal contribution. $^{\dag}$Project leader. $^{\diamond}$Corresponding authors.} 
\footnotetext[2]{\hspace{-1.5mm}$^{1}$AgiBot. $^{2}$SJTU. $^{3}$MMLab-CUHK. $^{4}$HIT.} 

\begin{abstract}
Recent advances in creative AI have enabled the synthesis of high-fidelity images and videos conditioned on language instructions. Building on these developments, text-to-video diffusion models have evolved into embodied world models (EWMs) capable of generating physically plausible scenes from language commands, effectively bridging vision and action in embodied AI applications. This work addresses the critical challenge of evaluating EWMs beyond general perceptual metrics to ensure the generation of physically grounded and action-consistent behaviors. We propose the Embodied World Model Benchmark (\Ours), a dedicated framework designed to evaluate EWMs based on three key aspects: visual scene consistency, motion correctness, and semantic alignment. Our approach leverages a meticulously curated dataset encompassing diverse scenes and motion patterns, alongside a comprehensive multi-dimensional evaluation toolkit, to assess and compare candidate models. The proposed benchmark not only identifies the limitations of existing video generation models in meeting the unique requirements of embodied tasks but also provides valuable insights to guide future advancements in the field. The dataset and evaluation tools are publicly available at \href{https://github.com/AgibotTech/EWMBench}{https://github.com/AgibotTech/EWMBench}.

\end{abstract}
\section{Introduction}
\label{sec:intro}

\begin{figure}[h]
    \centering
    \includegraphics[width=1\linewidth]{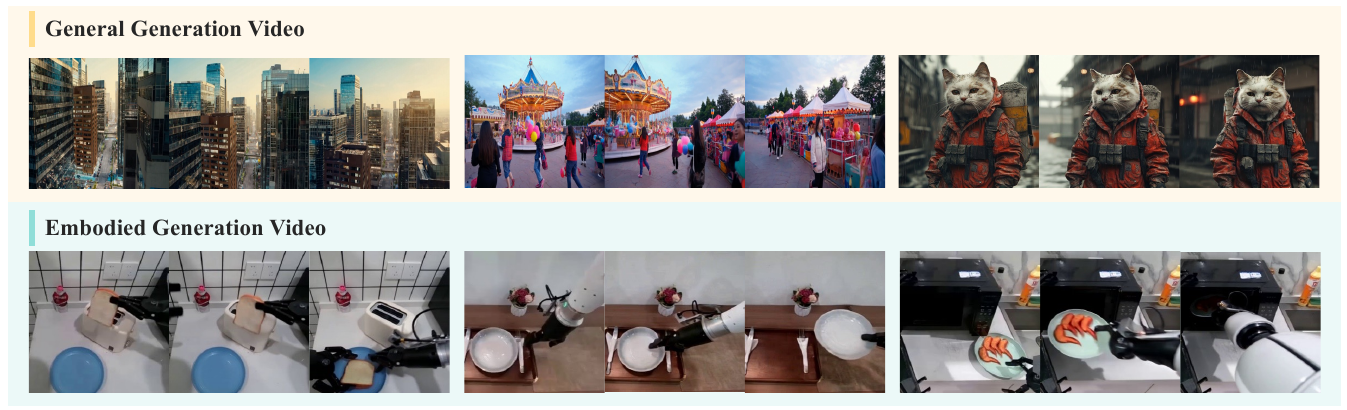}
    \caption{Comparison between general video generation and embodied video generation. Unlike general videos, embodied videos typically feature more structured scenes, consistent motion patterns, and clearer task logic.}
    \label{fig:general&embodied}
\end{figure}

Creative AI has advanced rapidly in recent years, propelled by innovations in model architectures—such as variational autoencoders (VAEs) and diffusion models—increased parameter scaling, and the availability of large-scale, high-quality datasets. These developments have empowered generative models to synthesize images and videos conditioned on language instructions with unprecedented fidelity and controllability. Building on the momentum of text-to-video diffusion models, recent efforts have expanded their scope from generating high-fidelity, high-resolution videos to serving as embodied world models~(EWMs) capable of synthesizing physically actionable scenes from language instructions (\emph{e.g.}, “move the robot arm approaching the cup”) or physical action instructions, \emph{i.e.}, an action policy sequence. This emerging capability establishes a critical link between vision and action in embodied AI, facilitating applications such as robotic manipulation, where instruction-conditioned trajectories must conform to physical and kinematic constraints.

Despite advancements in EWMs, a fundamental question remains unresolved: \emph{“How can we determine whether a video generation model qualifies as a good embodied world model, beyond merely serving as a general-purpose video generator?”} Addressing this question is essential for guiding model development and assessing a model’s ability to produce physically grounded, action-consistent behaviors. While existing video generation benchmarks~\cite{huang2024vbench} focus on perceptual metrics like visual fidelity, language alignment, and human preference, these criteria are insufficient for evaluating EWMs. Embodied generation tasks have unique requirements, such as coherence in embodiment motion and plausibility in action execution, as illustrated in Figure~\ref{fig:general&embodied}. For instance, in robotic manipulation scenarios, the background, object configuration, and embodiment structure (\emph{e.g.}, robot morphology) are expected to remain static, while only the robot's pose and interactions evolve according to instructions. This structured realism sets EWMs apart from general video generators and demands evaluation beyond conventional criteria.

In this work, we introduce a dedicated benchmark, Embodied World Model Benchmark (\Ours), to systematically assess embodiment motion fidelity and spatiotemporal consistency in robotic manipulation. We first formalize the benchmarking setup for EWMs. Given an initial video segment that specifies the embodiment (\emph{e.g.}, a robotic arm) and the environment, along with a manipulation instruction, the candidate EWM is tasked with autoregressively generating future frames depicting the embodiment’s motion until the instruction is completed. We design an evaluation protocol based on three key aspects: (1) \textbf{Visual Scene Consistency}, ensuring static elements like the background, objects, and embodiment structure remain unchanged during motion; (2) \textbf{Motion Correctness}, requiring the generated embodiment trajectory to be coherent and aligned with the task objective; and (3) \textbf{Semantic Alignment and Diversity}, assessing the model’s alignment with linguistic instructions and its ability to generalize across diverse tasks. For these aspects, we develop systematic evaluation tools, including prompt engineering with video-based MLLMs.

To benchmark EWMs under our proposed criteria, we construct a comprehensive colosseum consisting of a curated benchmark dataset and open-source evaluation tools. The dataset is built on Agibot-World~\cite{AgiBot2024agibotworld}, the largest real-world robotic manipulation dataset, featuring diverse tasks at scale. We select 30 candidate samples across ten tasks with clear action-ordering constraints, where correct execution requires understanding logical dependencies and affordances—posing significant challenges for embodied video generation. For each sample, static initial frames are clipped to ensure subsequent frames strictly reflect annotated language instructions without redundant movements. To reflect task diversity, we account for cases where multiple trajectories achieve the same goal and incorporate voxelized scoring to encourage variation. With this dataset, initial frames and language instructions are fed into different video generators, and the generated videos are compared against ground truth (GT) using various metrics.

\textbf{Contributions:} We summarize our contributions as follows: (1) We propose the first world generation benchmark tailored for embodied tasks, \Ours. (2)We curate a high-quality, diverse dataset for our benchmark evaluation. (3) We introduce and open-source the systematical evaluation metrics, which covers key aspects in the embodied world model generation. (4) we provide insights into the performance of existing video models on embodied generation tasks.

\newcommand{\cmark}{\textcolor{green!60!black}{\ding{51}}}
\newcommand{\xmark}{\textcolor{red}{\ding{55}}}


\section{Related Works}
\label{sec:related_works}

\subsection{Video Generative Models}

\noindent Diffusion-based video generation models have made significant advances in recent years, particularly in text-to-image (T2I) generation~\cite{blattmann2023stable, ho2020denoising, song2020denoising}and text-to-video (T2V) ~\cite{2023i2vgenxl, chen2023videocrafter1, ren2024consisti2v, guo2023animatediff, xing2025dynamicrafter} generation.  Recent works~\cite{kong2024hunyuanvideo, opensora, bao2024vidu} have explored the usage of Diffusion Transformers in the denoising process. Some researchers~\cite{cheang2024gr} in robot learning aim to leverage this knowledge to address the challenge of data scarcity in robotic data collection. In some works~\cite{brooks2024video, yang2023learning}video models are employed to predict and simulate future states of dynamic systems. In robotics, video models have been utilized to predict future frames based on textual and visual inputs~\cite{zhou2024robodreamer, huang2025enerverse, chi2024eva}.

\subsection{Evaluation of Video Generative Models}

\begin{table}[ht]
\centering
\scriptsize
\setlength{\tabcolsep}{4pt}
\resizebox{\textwidth}{!}{%
\begin{tabular}{l cc cccccc}
\toprule
\multirow{2}{*}{\textbf{Benchmark}} 
& \multicolumn{2}{c}{\textbf{Motion}} 
& \multirow{1}{*}{\textbf{Scene}} 
& \multicolumn{5}{c}{\textbf{Semantic}} \\
\cmidrule(lr){2-3} \cmidrule(lr){4-4} \cmidrule(lr){5-9}
& TrajD & TrajP & SceneC 
& DivM & TempC & InterL & ImgC & SemS \\
\midrule
TC-Bench~\cite{feng2024tc}                & \xmark & \xmark & \cmark & \cmark & \cmark & \cmark & \cmark & \xmark \\
Physics-IQ~\cite{motamed2025generative}  & \cmark & \cmark & \xmark & \xmark & \cmark & \cmark & \cmark & \xmark \\
VBench~\cite{huang2024vbench}            & \xmark & \xmark & \cmark & \xmark & \xmark & \xmark & \xmark & \cmark \\
VBench++~\cite{huang2024vbench++}        & \xmark & \xmark & \cmark & \xmark & \xmark & \xmark & \cmark & \cmark \\
PhyGenBench~\cite{phygenbench}           & \xmark & \xmark & \xmark & \xmark & \xmark & \xmark & \xmark & \cmark \\
T2V-CompBench~\cite{sun2024t2v}          & \xmark & \cmark & \xmark & \cmark & \xmark & \cmark & \xmark & \cmark \\
VMBench~\cite{ling2025vmbench}           & \cmark & \xmark & \cmark & \xmark & \xmark & \xmark & \cmark & \xmark \\
EvalCrafter~\cite{liu2023evalcrafter}    & \xmark & \xmark & \cmark & \cmark & \cmark & \cmark & \xmark & \cmark \\
T2VBench~\cite{ji2024t2vbench}           & \xmark & \xmark & \cmark & \cmark & \cmark & \cmark & \xmark & \xmark \\
\textit{EVA}~\cite{chi2024eva}           & \xmark & \cmark & \cmark & \xmark & \cmark & \cmark & \cmark & \cmark \\
\textbf{Ours}                             & \cmark & \cmark & \cmark & \cmark & \cmark & \cmark & \cmark & \cmark \\
\bottomrule
\end{tabular}
}
\vspace{0.5em}
\caption{Comparison of video generation benchmarks on 8 key evaluation dimensions: trajectory dynamics (TrajD), trajectory plausibility (TrajP), scene consistency (SceneC), diversity (DivM), temporal causality (TempC), interaction logic (InterL), image quality (ImgC), and semantic similarity (SemS).}
\label{tab:comparison_benchmark}
\end{table}

\noindent  With the rapid development of video generation models and their broad applications across various domains, evaluating these models has become increasingly important. Earlier approaches primarily relied on conventional metrics such as Fréchet Inception Distance (FID)\cite{heusel2017fid}, Inception Score (IS)\cite{salimans2016inceptionscore}, and Fréchet Video Distance (FVD)~\cite{unterthiner2019fvd}. However, these metrics primarily focus on visual appearance and offer limited insights into the diverse and complex capabilities of modern video generation models. Recent evaluation frameworks~\cite{huang2024vbench, huang2024vbench++} introduced a more structured approach that considers multiple capability dimensions. Nevertheless, these benchmarks still emphasize visual quality, including aesthetic appeal and motion smoothness. To address these limitations, specialized benchmarks have emerged. For instance, PhyGenBench~\cite{phygenbench} evaluates a model's understanding of physical laws using Vision-Language Models (VLMs), while T2V-CompBench~\cite{sun2024t2v} assesses compositionality, covering motion, actions, spatial relationships, and attributes.

  Although these efforts have significantly expanded evaluation dimensions, they remain focused on general video generation. In the context of world models, particularly in the embodied domain, video generation requires enhanced controllability, physical plausibility, and robust object interactions. However, previous works have generally overlooked critical factors such as action plausibility, object interactions, and manipulation. To bridge this gap, we propose \Ours, a systematic framework for evaluating embodied world models. A detailed comparison with existing benchmarks is provided in Table~\ref{tab:comparison_benchmark}.


\section{The \Ours Benchmark}
\label{sec:method}


\begin{figure}[h]
    \centering
    \includegraphics[width=0.95\linewidth]{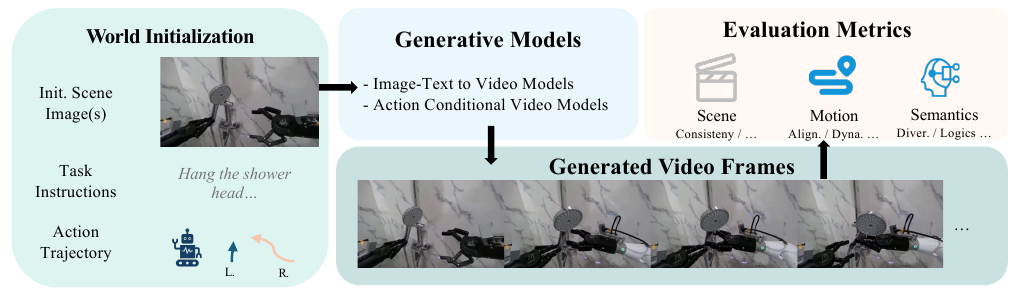}
    
    \caption{Overview of the \Ours benchmark design. The framework begins with unified world initialization, where generative models are instructed to produce predictive video frames based on initial scene images, task instructions, and optional action trajectories. These generated video frames are subsequently evaluated using multi-dimensional metrics, focusing on scene consistency, motion dynamics, and semantic alignment.}
    \label{fig:overall_pipeline}
    \vspace{-0.2cm}
    
\end{figure}

\subsection{\Ours Overview}
\noindent \textbf{Overview} The primary objective of this work is to establish a comprehensive evaluation benchmark for embodied world model generation. \Ours introduces three core components: (1) a unified world initialization, (2) a meticulously curated dataset for embodied manipulation tasks, and (3) systematic evaluation metrics. The complete pipeline is illustrated in Figure~\ref{fig:overall_pipeline}. The world is initialized using initial scene images and corresponding task instructions, with the sampled action trajectory being optional for generative models. Leveraging these unified input modalities, various generative models are instructed to produce video frames, while some contextual modalities may remain optional depending on the model. The outputs generated are evaluated using \Ours metrics, which focus on three critical factors: scene, motion, and semantics. These factors collectively form the foundation for evaluating robotic tasks. In this work, we primarily focus on robotic manipulation tasks, which are the most dominant and representative within the domain of embodied tasks. We plan to extend this framework to broader tasks in future work.

\noindent \textbf{Evaluation Task Formulation} An embodied world model generates a video as expressed in Equation~\ref{equ:overall_process}, where $\mathcal{I}$, $\mathcal{L}$, and $\mathcal{T}$ represent the input context image, language, and trajectory, respectively. We provide this unified information, including up to four initial images. The action trajectory, formatted as a sequence of 6D poses, is optional for generation model inference. The function $f_\text{proc}$ represents model-specific preprocessing. To ensure fairness, we apply $v_\text{norm}$ to normalize the raw generated video frames before evaluation. The evaluation framework comprises a multi-modal LLM for high-level semantic analysis, a trajectory detector for low-level trajectory-based evaluation, and several visual foundation models for visual feature processing.

\begin{equation}
\label{equ:overall_process}
\mathbf{V} = v_\text{norm}(g_\text{world}(f_\text{proc}(\mathcal{I, L, T})))
\end{equation}

\subsection{Dataset Construction}



We developed our evaluation dataset using the open-source Agibot-World dataset. Ten tasks were carefully selected based on their clear operational goals and sequential dependencies, covering both household and industrial contexts. These tasks emphasize action-ordering constraints that require reasoning about affordances and procedural contexts. To ensure diverse motion patterns, action trajectories were encoded into voxel grids, and a greedy algorithm was employed to select the most diverse trajectories for each task. An overview of the constructed dataset is in Figure~\ref{fig:constructed_dataset}. Details on task descriptions can be found in the Appendix.

For fine-grained evaluation, we adopted a task-oriented decomposition strategy. Each high-level task was broken down into a sequence of 4 to 10 atomic sub-actions, with each sub-action paired with a step-level caption. This approach guarantees a one-to-one alignment between video segments, sub-action labels, and their corresponding linguistic descriptions.

\begin{figure}
    \centering
    \includegraphics[width=1\linewidth]{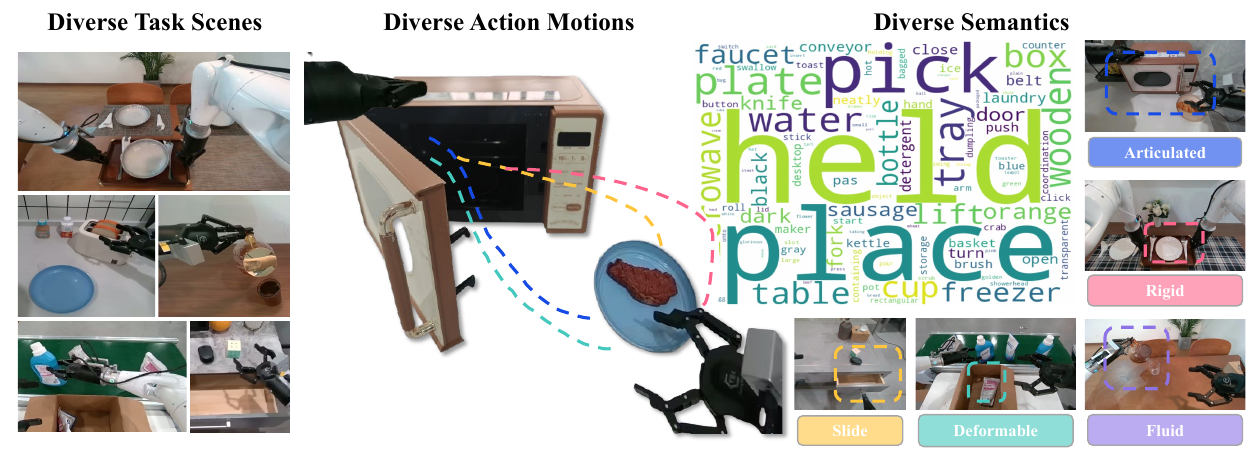}
    \vspace{0.2em}
    \caption{Overview of the constructed dataset. \textbf{Left}: Task scenes spanning household, commercial, and industrial environments. \textbf{Middle}: Diverse task-specific trajectory variations within each scene. \textbf{Right}: Broad semantic coverage across various manipulation contexts.}
    \label{fig:constructed_dataset}
\end{figure}

\subsection{Evaluation Metrics}

\Ours systematically evaluates three dimensions to ensure the generated outputs are visually realistic, action plausible, and semantically meaningful. We leave the metric definition details in the Appendix.

\noindent \textbf{A. Scene.~} We introduce the \underline{Scene Consistency} metric, which examines visual layouts, object permanence, and viewpoint coherence. DINOv2, fine-tuned on an embodied dataset, extracts patch-level frame representations. Cosine similarity between patch embeddings of consecutive and initial frames quantifies frame-to-frame consistency. Higher scores indicate stable scene structures and coherent viewpoints throughout the video.

\noindent \textbf{B. Action Motion.~} The quality of generated motions is evaluated through a \textbf{Trajectory-Based} Evaluation, which compares generated trajectories with ground truth trajectories. The trajectories capture physical consistency, task logic, and interaction constraints. In our setup, we use the end-effector's (EEF) trajectory as the evaluation target and EEF is detected with our finetuned detector. \underline{Symmetric Hausdorff Distance} (HSD) measures spatial alignment by calculating the maximum deviation between points on generated and GT trajectories. \noindent \underline{Normalized Dynamic Time Warping} (NDTW) captures spatial-temporal alignment, ensuring correct sequence and timing of motions. \underline{Dynamic Consistency} (DYN) evaluates motion dynamics, such as velocity and acceleration, using Wasserstein distance with motion normalization. To ensure fairness, generative models are required to produce three candidate trajectories for each task. The best trajectory is selected based on Hausdorff distance.

\noindent \textbf{C. Semantics.~} Semantic evaluation focuses on (1) alignment between task instructions and generated videos and (2) diversity within the task space. For \underline{semantic alignment}, we use the generated video's language caption as an intermediate representation, comparing them to ground truth annotations to compute an alignment score. Captions are extracted at three levels, with details on the prompt design provided in Section~\ref{sec:prompt_suite}. For \underline{semantic diversity}, we use CLIP model, global video features are extracted, and the diversity score is computed as $1 - \text{similarity}$. This reflects the model’s ability to generalize and produce varied outputs.


\subsection{MLLM Prompt Suite Design}
\label{sec:prompt_suite}
\Ours Prompt suite is designed to be compact yet representative. We perform this evaluation across three levels of language analysis.  Full prompts are in the project page.

\noindent \textbf{Global Video Caption Representation}:  
At the global level, a video MLLM generates a compact caption summarizing the entire video. This caption is compared with the raw task instruction to evaluate overall alignment between the task goal and the generated video’s content using BLEU score.

\noindent \textbf{Key Steps Description}:  
Robot tasks often involve multiple key steps that may be lost in global representations. To address this, the video MLLM produces a detailed, step-by-step description of the task’s key steps. These descriptions are compared with GT step descriptions generated by the MLLM using CLIP score.

\noindent \textbf{Logical Error Punishment}:  
Logical errors, such as hallucinations or spatial inconsistencies, are critical in robotic applications as they can lead to unsafe outcomes. The MLLM evaluates generated videos for commonsense violations, explicitly penalizing errors like hallucinated object manipulations or illogical spatial relationships. These penalties ensure that the model prioritizes realistic and coherent task execution.

\vspace{-0.2cm}
\section{Experiments}
\label{sec:experiments}
\vspace{-0.3cm}
\noindent \textbf{Models} We evaluate \Ours across seven video generation models categorized as open-source, commercial, and domain-adapted. The open-source models include OpenSora 2.0\cite{peng2025open}, which demonstrates strong performance on VBench with low training costs; LTX\cite{hacohen2024ltx}, capable of real-time generation for interactive tasks; and COSMOS-7B\cite{agarwal2025cosmos}, pre-trained for digital twin applications. Commercial models optimized for zero-shot generation include Kling-1.6\cite{Kuaishou2025Kling} and Hailuo I2V-01-live\cite{HailuoAI2025Hailuo}, both of which rank highly in recent benchmarks. Domain-adapted models fine-tuned for embodied scene understanding and action prediction include LTX\_FT, a fine-tuned version of LTX, and EnerVerse, which is specifically designed for embodied scenarios. At present, we primarily focus on the Image-Text-to-Video setting, as no open-source action-conditioned video generation models are currently available. The evaluation of such models is left for future work. Nonetheless, with our unified input format and visual-space evaluation operations, \Ours could also support this evaluation in the future. For the current evaluation, we tested 10 tasks, each consisting of 10 ground-truth episodes. Three videos were generated per model per episode, and the best prediction was selected using a best-of-three strategy, resulting in a total of 2,100 videos.

\vspace{-0.2cm}
\subsection{Evaluation Results}
\vspace{-0.2cm}

We evaluate models across dimensions using normalized scores between 0 and 1, where higher values indicate better performance. Results in Table~\ref{tab:wmbm_scores_bolded} show that domain-adapted models, such as EnerVerse and LTX\_FT, consistently outperform commercial models (e.g., Kling, Hailuo) and open-source models (e.g., COSMOS, OpenSora, LTX). This highlights the effectiveness of domain-specific fine-tuning in capturing motion dynamics and task semantics. Notably, EnerVerse and Kling demonstrate strong semantic alignment, reflecting a solid understanding of task logic.

\begin{figure}
    \centering
    \includegraphics[width=1\linewidth]{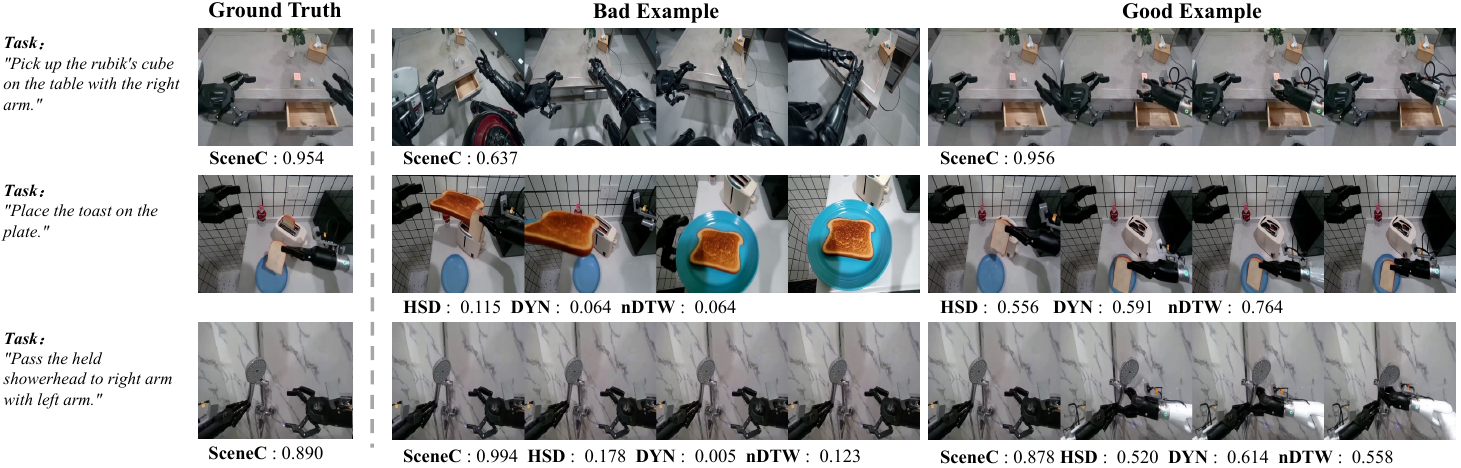}
    \vspace{0.5em}
    \caption{\textbf{Typical examples.} \Ours scores align well with scene and motion accuracy, demonstrating the interpretability and robustness of the proposed metrics.}
    \label{fig:good_bad case}
\end{figure}

To validate the reliability of our evaluation, Figure~\ref{fig:good_bad case} provides representative examples. Low scene consistency is marked by changes in spatial layout and object presence, while high scene consistency preserves both. Poor trajectory consistency features mismatched end-effector motion and task failure, whereas good cases exhibit motion patterns closely aligned with the ground truth, ensuring task success. Importantly, scene and trajectory consistency are complementary: visually plausible but static videos may score high in scene consistency while lacking meaningful motion. This emphasizes the need for a systematic, multi-dimensional evaluation approach.

\begin{figure}
    \centering
    \includegraphics[width=0.9\linewidth]{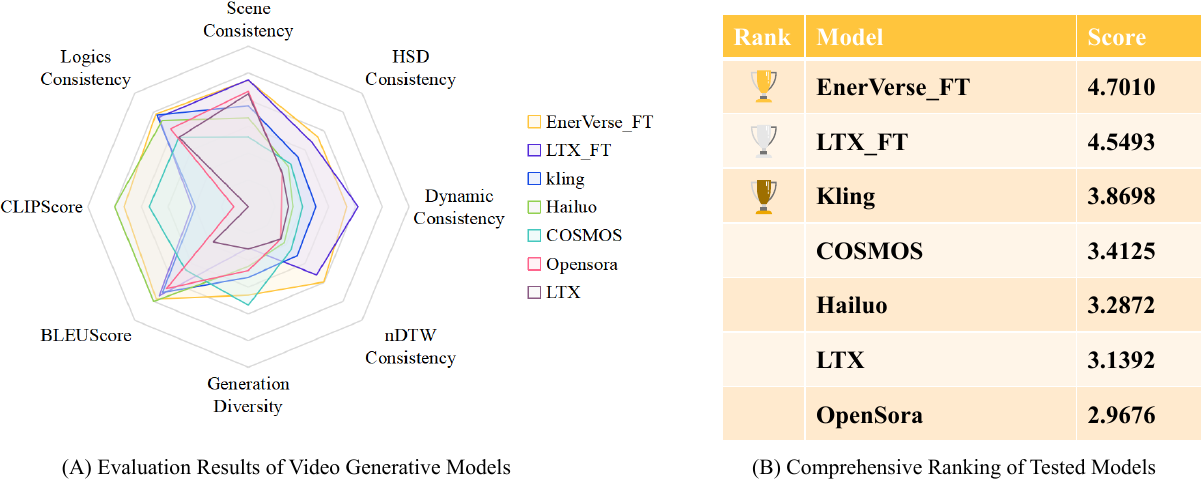}
    \caption{Evaluation Results of Video Generative Models. }
    \label{fig:radar chart}
\end{figure}

\begin{table*}[htbp]
\centering
\resizebox{\textwidth}{!}{
\begin{tabular}{llcccccccccccc}
\toprule
\multirow{2}{*}{Type} & \multirow{2}{*}{Model} & Scene & \multicolumn{4}{c}{Motion} & \multicolumn{5}{c}{Semantics} & \multirow{2}{*}{Overall} \\
\cmidrule(r){3-3} \cmidrule(r){4-7} \cmidrule(r){8-12}
& & SceneC & HSD & Dyn & nDTW & Avg. & Diversity & BLEU & CLIP & Logics & Avg. & \\
\midrule
\multirow{2}{*}{Dom.} 
& EnerVerse\_FT\cite{huang2025enerverse} & 0.9427 & \textbf{0.5356} & 0.5363 & \textbf{0.5957} & \textbf{1.6676} & 0.0691 & \textbf{0.1800} & 0.8638 & \textbf{0.9778} & \textbf{2.0907} & \textbf{4.7010} \\
& LTX\_FT  & \textbf{0.9436} & 0.4758 & \textbf{0.6197} & 0.5208 & 1.6163 & 0.0162 & 0.1740 & 0.8548 & 0.9444 & 1.9894 & 4.5493 \\
\midrule
\multirow{2}{*}{Comm.} 
& Kling\cite{Kuaishou2025Kling} & 0.8888 & 0.3231 & 0.3047 & 0.3162 & 0.9440 & 0.0493 & 0.1675 & 0.8535 & 0.9667 & 2.0370 & 3.8698 \\
& Hailuo\cite{HailuoAI2025Hailuo} & 0.8577 & 0.2229 & 0.1344 & 0.1789 & 0.5362 & 0.0370 & 0.1848 & \textbf{0.8857} & 0.9111 & 2.0186 & 3.4125 \\
\midrule
\multirow{3}{*}{Open.} 
& COSMOS\cite{agarwal2025cosmos} & 0.7963 & 0.2500 & 0.2052 & 0.2533 & 0.7085 & \textbf{0.0803} & 0.1230 & 0.8458 & 0.7333 & 1.7824 & 3.2872 \\
& OpenSora\cite{peng2025open} & 0.9210 & 0.1548 & 0.0474 & 0.1420 & 0.3442 & 0.0415 & 0.1598 & 0.8505 & 0.8222 & 1.8739 & 3.1392 \\
& LTX\cite{hacohen2024ltx} & 0.9156 & 0.1575 & 0.1002 & 0.1425 & 0.4002 & 0.0174 & 0.0687 & 0.8324 & 0.7333 & 1.6518 & 2.9676 \\
\bottomrule
\end{tabular}
}
\vspace{0.5em}

\caption{Evaluation results categorized into task scene, action motion, and semantics. }
\label{tab:wmbm_scores_bolded}
\end{table*}

\subsection{Human Evaluation}

To evaluate the alignment between automated metrics and human judgment, we conducted a human evaluation on videos generated by four representative models: LTX\_FT, Kling-1.6, Hailuo I2V-01-live, and OpenSora-2.0. Annotators ranked the predictions based on overall quality, assigning 3 points to the best, 2 to the second-best, and 0 to the worst. Final rankings were derived by aggregating scores across all annotators and samples, with multiple review rounds ensuring annotation reliability.

\begin{figure}
    \centering
    \includegraphics[width=1\linewidth]{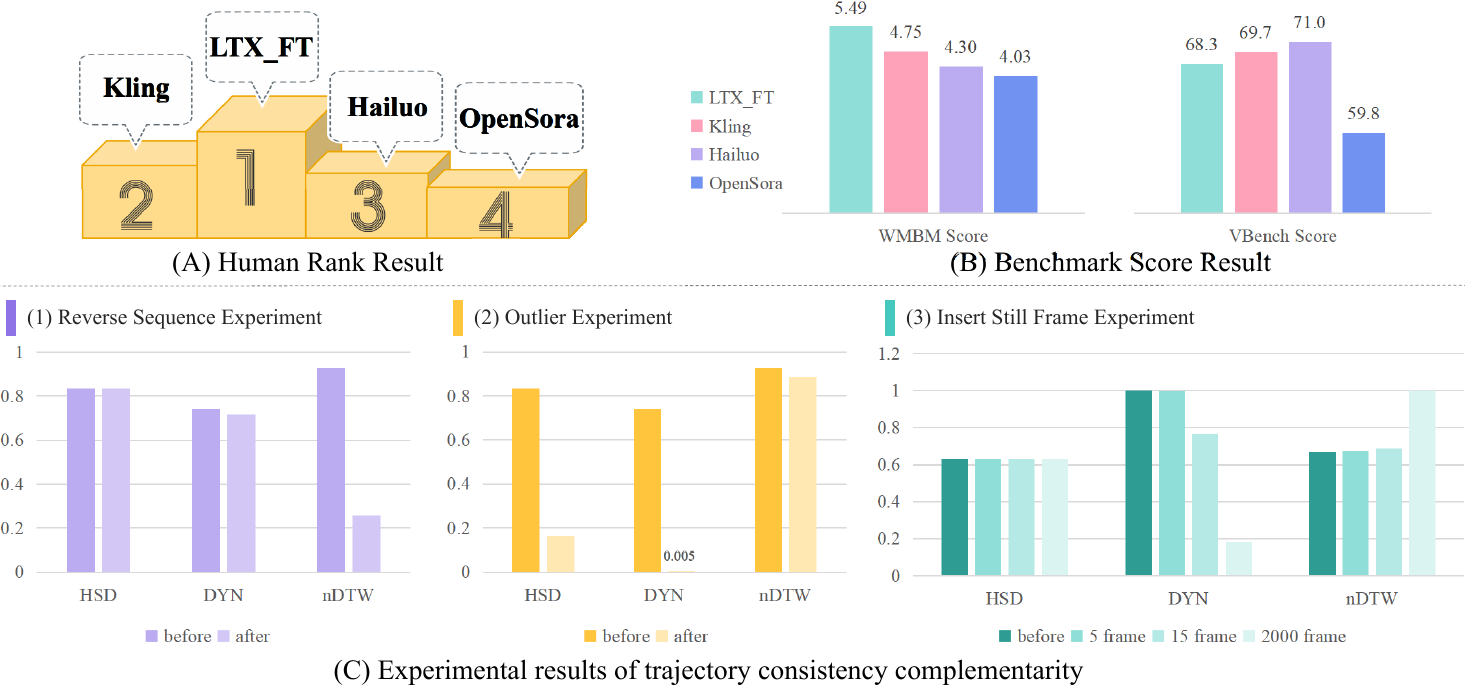}
    \caption{(A) Aggregated human rankings of model predictions. (B) Comparison of rankings produced by \Ours and VBench, highlighting \Ours's closer alignment with human judgments.(C) Complementarity of trajectory metrics.}
\vspace{-0.2cm}
    \label{fig:bar chart}
\end{figure}

We then compared this aggregated human rankings (Figure~\ref{fig:bar chart} (A)) with those produced by \Ours and VBench~\cite{huang2024vbench}, one of the most popular video  generation evaluation benchmarks. As shown in Figure~\ref{fig:bar chart} (B), \Ours's rankings align more closely with human judgments than VBench rankings, indicating stronger consistency with human perception.

\subsection{Complementarity of Trajectory Metrics}

To validate the necessity of employing all three trajectory consistency metrics—HSD, nDTW, and DYN—we conducted controlled experiments involving sequence reversal, outlier insertion, and frame repetition. As shown in Figure~\ref{fig:bar chart} (C), each metric responds uniquely, demonstrating its specific strengths. In the sequence reversal test, only nDTW showed a significant drop due to its sensitivity to temporal order, underscoring its role in detecting alignment errors. In the outlier test, HSD and DYN experienced substantial declines, reflecting their focus on spatial accuracy and motion integrity—both essential for safe and precise embodied execution. In the frame repetition test, nDTW increased due to repeated alignment, while DYN decreased, highlighting its sensitivity to motion smoothness. These findings confirm the complementary roles of the three metrics in providing a comprehensive evaluation of trajectory quality.

\subsection{Further Analysis and Discussions}

\noindent \textbf{Characteristics of SOTA Models}
We present key findings from \Ours’s evaluation, focusing on trade-offs, model characteristics, and the impact of domain adaptation in embodied video generation. Qualitative examples illustrating these failure modes are provided in Appendix A.4.

\noindent \underline{Domain-adapted models} show the best overall performance, particularly in semantic and dynamic dimensions, demonstrating that targeted fine-tuning significantly enhances task understanding and motion alignment. However, they occasionally exhibit empty grasping behaviors, revealing limitations in fine-grained action grounding.

\noindent \underline{Kling} achieves the best overall performance among general commercial and open-source video models, demonstrating strong and robust capabilities.

\noindent \underline{Hailuo} performs reasonably well in zero-shot embodied scenarios, but its generated scenes often appear cartoon-like, limiting visual realism.

\noindent \underline{COSMOS} and \noindent \underline{LTX} display a bias toward human hand representations and frequently fail to adapt semantic understanding to robotic contexts. LTX, in particular, suffers from abrupt scene transitions, inconsistent task execution, and and a tendency to generate stationary states in action sequences. In contrast, COSMOS struggles to maintain consistent viewpoints, highlighting inadequate control of camera parameters.

\noindent \underline{OpenSora} shows partial understanding of task scenes, action motions, and semantic alignment in manipulation tasks. However, it suffers from jittery robotic arm movements and frequently generates static videos.

\noindent \textbf{Comparison with VBench Metrics} Our experiments reveal that VBench struggles to separate foreground and background features, limiting the effectiveness of its subject-level metrics. In contrast, our Scene Consistency metric, leveraging fine-tuned DINOv2, excels at capturing layout structure and is more sensitive to viewpoint changes. This heightened sensitivity enables the detection of visual instability, a critical factor in embodied video generation. Additional details and feature map visualizations are provided in the Appendix.
\vspace{-0.2cm}
\section{Conclusions and Limitations}
\vspace{-0.3cm}

\label{sec:discussion}
 In this work, we propose \Ours, a comprehensive benchmark suite for evaluating embodied world generation models. With its multi-dimensional, human-aligned metric design and a motion-diverse, multi-scene dataset, \Ours serves as a valuable tool for measuring progress in embodied world model development.

\noindent \textbf{Limitations and Future Work.}  First, our method currently focuses on the trajectory of the robotic arm's end-effector, but future work will incorporate the state and configuration of the entire arm. Second, the current evaluation is conducted in fixed-viewpoint scenes; future research will explore flexible viewpoints, such as dynamic camera setups. Lastly, we aim to extend the scope of embodied tasks—from the current manipulation tasks to more diverse domains, including navigation and mobile manipulation.

{
    \small
    \bibliographystyle{plainnat}
    \bibliography{main}

\begin{thebibliography}{43}
\providecommand{\natexlab}[1]{#1}
\providecommand{\url}[1]{\texttt{#1}}
\expandafter\ifx\csname urlstyle\endcsname\relax
  \providecommand{\doi}[1]{doi: #1}\else
  \providecommand{\doi}{doi: \begingroup \urlstyle{rm}\Url}\fi

\bibitem[Agarwal et~al.(2025)Agarwal, Ali, Bala, Balaji, Barker, Cai, Chattopadhyay, Chen, Cui, Ding, et~al.]{agarwal2025cosmos}
Niket Agarwal, Arslan Ali, Maciej Bala, Yogesh Balaji, Erik Barker, Tiffany Cai, Prithvijit Chattopadhyay, Yongxin Chen, Yin Cui, Yifan Ding, et~al.
\newblock Cosmos world foundation model platform for physical ai.
\newblock \emph{arXiv preprint arXiv:2501.03575}, 2025.

\bibitem[AgiBot(2024)]{AgiBot2024agibotworld}
AgiBot.
\newblock Agibot world.
\newblock \url{https://agibot-world.com}, 2024.

\bibitem[Aharon et~al.(2022)Aharon, Orfaig, and Bobrovsky]{aharon2022bot}
Nir Aharon, Roy Orfaig, and Ben-Zion Bobrovsky.
\newblock Bot-sort: Robust associations multi-pedestrian tracking.
\newblock \emph{arXiv preprint arXiv:2206.14651}, 2022.

\bibitem[Bao et~al.(2024)Bao, Xiang, Yue, He, Zhu, Zheng, Zhao, Liu, Wang, and Zhu]{bao2024vidu}
Fan Bao, Chendong Xiang, Gang Yue, Guande He, Hongzhou Zhu, Kaiwen Zheng, Min Zhao, Shilong Liu, Yaole Wang, and Jun Zhu.
\newblock Vidu: a highly consistent, dynamic and skilled text-to-video generator with diffusion models.
\newblock \emph{arXiv preprint arXiv:2405.04233}, 2024.

\bibitem[Blattmann et~al.(2023)Blattmann, Dockhorn, Kulal, Mendelevitch, Kilian, Lorenz, Levi, English, Voleti, Letts, et~al.]{blattmann2023stable}
Andreas Blattmann, Tim Dockhorn, Sumith Kulal, Daniel Mendelevitch, Maciej Kilian, Dominik Lorenz, Yam Levi, Zion English, Vikram Voleti, Adam Letts, et~al.
\newblock Stable video diffusion: Scaling latent video diffusion models to large datasets.
\newblock \emph{arXiv preprint arXiv:2311.15127}, 2023.

\bibitem[Brooks et~al.(2024)Brooks, Peebles, Holmes, DePue, Guo, Jing, Schnurr, Taylor, Luhman, Luhman, et~al.]{brooks2024video}
Tim Brooks, Bill Peebles, Connor Holmes, Will DePue, Yufei Guo, Li~Jing, David Schnurr, Joe Taylor, Troy Luhman, Eric Luhman, et~al.
\newblock Video generation models as world simulators. 2024.
\newblock \emph{URL https://openai. com/research/video-generation-models-as-world-simulators}, 3:\penalty0 1, 2024.

\bibitem[Cheang et~al.(2024)Cheang, Chen, Jing, Kong, Li, Li, Liu, Wu, Xu, Yang, et~al.]{cheang2024gr}
Chi-Lam Cheang, Guangzeng Chen, Ya~Jing, Tao Kong, Hang Li, Yifeng Li, Yuxiao Liu, Hongtao Wu, Jiafeng Xu, Yichu Yang, et~al.
\newblock Gr-2: A generative video-language-action model with web-scale knowledge for robot manipulation.
\newblock \emph{arXiv preprint arXiv:2410.06158}, 2024.

\bibitem[Chen et~al.(2023)Chen, Xia, He, Zhang, Cun, Yang, Xing, Liu, Chen, Wang, Weng, and Shan]{chen2023videocrafter1}
Haoxin Chen, Menghan Xia, Yingqing He, Yong Zhang, Xiaodong Cun, Shaoshu Yang, Jinbo Xing, Yaofang Liu, Qifeng Chen, Xintao Wang, Chao Weng, and Ying Shan.
\newblock Videocrafter1: Open diffusion models for high-quality video generation, 2023.

\bibitem[Cheng et~al.(2024)Cheng, Song, Ge, Liu, Wang, and Shan]{cheng2024yolo}
Tianheng Cheng, Lin Song, Yixiao Ge, Wenyu Liu, Xinggang Wang, and Ying Shan.
\newblock Yolo-world: Real-time open-vocabulary object detection.
\newblock In \emph{Proceedings of the IEEE/CVF Conference on Computer Vision and Pattern Recognition}, pages 16901--16911, 2024.

\bibitem[Chi et~al.(2024)Chi, Zhang, Fan, Qi, Zhang, Chen, Chan, Xue, Luo, Zhang, et~al.]{chi2024eva}
Xiaowei Chi, Hengyuan Zhang, Chun-Kai Fan, Xingqun Qi, Rongyu Zhang, Anthony Chen, Chi-min Chan, Wei Xue, Wenhan Luo, Shanghang Zhang, et~al.
\newblock Eva: An embodied world model for future video anticipation.
\newblock \emph{arXiv preprint arXiv:2410.15461}, 2024.

\bibitem[Feng et~al.(2024)Feng, Li, Saxon, Fu, Chen, and Wang]{feng2024tc}
Weixi Feng, Jiachen Li, Michael Saxon, Tsu-jui Fu, Wenhu Chen, and William~Yang Wang.
\newblock Tc-bench: Benchmarking temporal compositionality in text-to-video and image-to-video generation.
\newblock \emph{arXiv preprint arXiv:2406.08656}, 2024.

\bibitem[Guo et~al.(2023)Guo, Yang, Rao, Liang, Wang, Qiao, Agrawala, Lin, and Dai]{guo2023animatediff}
Yuwei Guo, Ceyuan Yang, Anyi Rao, Zhengyang Liang, Yaohui Wang, Yu~Qiao, Maneesh Agrawala, Dahua Lin, and Bo~Dai.
\newblock Animatediff: Animate your personalized text-to-image diffusion models without specific tuning.
\newblock \emph{arXiv preprint arXiv:2307.04725}, 2023.

\bibitem[HaCohen et~al.(2024)HaCohen, Chiprut, Brazowski, Shalem, Moshe, Richardson, Levin, Shiran, Zabari, Gordon, et~al.]{hacohen2024ltx}
Yoav HaCohen, Nisan Chiprut, Benny Brazowski, Daniel Shalem, Dudu Moshe, Eitan Richardson, Eran Levin, Guy Shiran, Nir Zabari, Ori Gordon, et~al.
\newblock Ltx-video: Realtime video latent diffusion.
\newblock \emph{arXiv preprint arXiv:2501.00103}, 2024.

\bibitem[Hailuo(2025)]{HailuoAI2025Hailuo}
Hailuo.
\newblock Hailuoai.
\newblock \url{https://hailuoai.video/}, 2025.

\bibitem[Heusel et~al.(2017)Heusel, Ramsauer, Unterthiner, Nessler, and Hochreiter]{heusel2017fid}
Martin Heusel, Hubert Ramsauer, Thomas Unterthiner, Bernhard Nessler, and Sepp Hochreiter.
\newblock {GANs} trained by a two time-scale update rule converge to a local nash equilibrium.
\newblock In \emph{Advances in neural information processing systems}, 2017.

\bibitem[Ho et~al.(2020)Ho, Jain, and Abbeel]{ho2020denoising}
Jonathan Ho, Ajay Jain, and Pieter Abbeel.
\newblock Denoising diffusion probabilistic models.
\newblock \emph{Advances in neural information processing systems}, 33:\penalty0 6840--6851, 2020.

\bibitem[Huang et~al.(2025)Huang, Chen, Zhou, Chen, Jiang, Hu, Liao, Gao, Li, Yao, et~al.]{huang2025enerverse}
Siyuan Huang, Liliang Chen, Pengfei Zhou, Shengcong Chen, Zhengkai Jiang, Yue Hu, Yue Liao, Peng Gao, Hongsheng Li, Maoqing Yao, et~al.
\newblock Enerverse: Envisioning embodied future space for robotics manipulation.
\newblock \emph{arXiv preprint arXiv:2501.01895}, 2025.

\bibitem[Huang et~al.(2024{\natexlab{a}})Huang, He, Yu, Zhang, Si, Jiang, Zhang, Wu, Jin, Chanpaisit, et~al.]{huang2024vbench}
Ziqi Huang, Yinan He, Jiashuo Yu, Fan Zhang, Chenyang Si, Yuming Jiang, Yuanhan Zhang, Tianxing Wu, Qingyang Jin, Nattapol Chanpaisit, et~al.
\newblock Vbench: Comprehensive benchmark suite for video generative models.
\newblock In \emph{Proceedings of the IEEE/CVF Conference on Computer Vision and Pattern Recognition}, 2024{\natexlab{a}}.

\bibitem[Huang et~al.(2024{\natexlab{b}})Huang, Zhang, Xu, He, Yu, Dong, Ma, Chanpaisit, Si, Jiang, et~al.]{huang2024vbench++}
Ziqi Huang, Fan Zhang, Xiaojie Xu, Yinan He, Jiashuo Yu, Ziyue Dong, Qianli Ma, Nattapol Chanpaisit, Chenyang Si, Yuming Jiang, et~al.
\newblock Vbench++: Comprehensive and versatile benchmark suite for video generative models.
\newblock \emph{arXiv preprint arXiv:2411.13503}, 2024{\natexlab{b}}.

\bibitem[Ji et~al.(2024)Ji, Xiao, Tai, and Huo]{ji2024t2vbench}
Pengliang Ji, Chuyang Xiao, Huilin Tai, and Mingxiao Huo.
\newblock T2vbench: Benchmarking temporal dynamics for text-to-video generation.
\newblock In \emph{Proceedings of the IEEE/CVF Conference on Computer Vision and Pattern Recognition}, pages 5325--5335, 2024.

\bibitem[Kong et~al.(2024)Kong, Tian, Zhang, Min, Dai, Zhou, Xiong, Li, Wu, Zhang, et~al.]{kong2024hunyuanvideo}
Weijie Kong, Qi~Tian, Zijian Zhang, Rox Min, Zuozhuo Dai, Jin Zhou, Jiangfeng Xiong, Xin Li, Bo~Wu, Jianwei Zhang, et~al.
\newblock Hunyuanvideo: A systematic framework for large video generative models.
\newblock \emph{arXiv preprint arXiv:2412.03603}, 2024.

\bibitem[Kuaishou(2025)]{Kuaishou2025Kling}
Kuaishou.
\newblock Kling.
\newblock \url{https://app.klingai.com/cn/}, 2025.

\bibitem[Ling et~al.(2025)Ling, Zhu, Wu, Li, Feng, Yang, Hao, Zhu, Wu, and Chu]{ling2025vmbench}
Xinrang Ling, Chen Zhu, Meiqi Wu, Hangyu Li, Xiaokun Feng, Cundian Yang, Aiming Hao, Jiashu Zhu, Jiahong Wu, and Xiangxiang Chu.
\newblock Vmbench: A benchmark for perception-aligned video motion generation.
\newblock \emph{arXiv preprint arXiv:2503.10076}, 2025.

\bibitem[Liu et~al.(2024)Liu, Cun, Liu, Wang, Zhang, Chen, Liu, Zeng, Chan, and Shan]{liu2023evalcrafter}
Yaofang Liu, Xiaodong Cun, Xuebo Liu, Xintao Wang, Yong Zhang, Haoxin Chen, Yang Liu, Tieyong Zeng, Raymond Chan, and Ying Shan.
\newblock Evalcrafter: Benchmarking and evaluating large video generation models.
\newblock In \emph{Proceedings of the IEEE/CVF Conference on Computer Vision and Pattern Recognition}, 2024.

\bibitem[Meng et~al.(2024)Meng, Liao, Tan, Shao, Lu, Zhang, Cheng, Li, Qiao, and Luo]{phygenbench}
Fanqing Meng, Jiaqi Liao, Xinyu Tan, Wenqi Shao, Quanfeng Lu, Kaipeng Zhang, Yu~Cheng, Dianqi Li, Yu~Qiao, and Ping Luo.
\newblock Towards world simulator: Crafting physical commonsense-based benchmark for video generation.
\newblock \emph{arXiv preprint arXiv:2410.05363}, 2024.

\bibitem[Motamed et~al.(2025)Motamed, Culp, Swersky, Jaini, and Geirhos]{motamed2025generative}
Saman Motamed, Laura Culp, Kevin Swersky, Priyank Jaini, and Robert Geirhos.
\newblock Do generative video models learn physical principles from watching videos?
\newblock \emph{arXiv preprint arXiv:2501.09038}, 2025.

\bibitem[M{\"u}ller(2007)]{muller2007dynamic}
Meinard M{\"u}ller.
\newblock Dynamic time warping.
\newblock \emph{Information retrieval for music and motion}, pages 69--84, 2007.

\bibitem[Oquab et~al.(2023)Oquab, Darcet, Moutakanni, Vo, Szafraniec, Khalidov, Fernandez, Haziza, Massa, El-Nouby, et~al.]{oquab2023dinov2}
Maxime Oquab, Timoth{\'e}e Darcet, Th{\'e}o Moutakanni, Huy Vo, Marc Szafraniec, Vasil Khalidov, Pierre Fernandez, Daniel Haziza, Francisco Massa, Alaaeldin El-Nouby, et~al.
\newblock Dinov2: Learning robust visual features without supervision.
\newblock \emph{arXiv preprint arXiv:2304.07193}, 2023.

\bibitem[Peng et~al.(2025)Peng, Zheng, Shen, Young, Guo, Wang, Xu, Liu, Jiang, Li, et~al.]{peng2025open}
Xiangyu Peng, Zangwei Zheng, Chenhui Shen, Tom Young, Xinying Guo, Binluo Wang, Hang Xu, Hongxin Liu, Mingyan Jiang, Wenjun Li, et~al.
\newblock Open-sora 2.0: Training a commercial-level video generation model in $200 k$.
\newblock \emph{arXiv preprint arXiv:2503.09642}, 2025.

\bibitem[Ren et~al.(2024)Ren, Yang, Zhang, Wei, Du, Huang, and Chen]{ren2024consisti2v}
Weiming Ren, Harry Yang, Ge~Zhang, Cong Wei, Xinrun Du, Stephen Huang, and Wenhu Chen.
\newblock Consisti2v: Enhancing visual consistency for image-to-video generation.
\newblock \emph{arXiv preprint arXiv:2402.04324}, 2024.

\bibitem[Salimans et~al.(2016)Salimans, Goodfellow, Zaremba, Cheung, Radford, Chen, and Chen]{salimans2016inceptionscore}
Tim Salimans, Ian Goodfellow, Wojciech Zaremba, Vicki Cheung, Alec Radford, Xi~Chen, and Xi~Chen.
\newblock Improved techniques for training gans.
\newblock In \emph{Advances in neural information processing systems}, 2016.

\bibitem[Serra(1998)]{serra1998hausdorff}
Jean Serra.
\newblock Hausdorff distances and interpolations.
\newblock \emph{Computational Imaging and Vision}, 12:\penalty0 107--114, 1998.

\bibitem[Song et~al.(2020)Song, Meng, and Ermon]{song2020denoising}
Jiaming Song, Chenlin Meng, and Stefano Ermon.
\newblock Denoising diffusion implicit models.
\newblock \emph{arXiv preprint arXiv:2010.02502}, 2020.

\bibitem[Sun et~al.(2024)Sun, Huang, Liu, Wu, Xu, Li, and Liu]{sun2024t2v}
Kaiyue Sun, Kaiyi Huang, Xian Liu, Yue Wu, Zihan Xu, Zhenguo Li, and Xihui Liu.
\newblock T2v-compbench: A comprehensive benchmark for compositional text-to-video generation.
\newblock \emph{arXiv preprint arXiv:2407.14505}, 2024.

\bibitem[Unterthiner et~al.(2019)Unterthiner, van Steenkiste, Kurach, Marinier, Michalski, and Gelly]{unterthiner2019fvd}
Thomas Unterthiner, Sjoerd van Steenkiste, Karol Kurach, Rapha{\"e}l Marinier, Marcin Michalski, and Sylvain Gelly.
\newblock {FVD}: A new metric for video generation.
\newblock In \emph{ICLRW}, 2019.

\bibitem[Villani and Villani(2009)]{villani2009wasserstein}
C{\'e}dric Villani and C{\'e}dric Villani.
\newblock The wasserstein distances.
\newblock \emph{Optimal transport: old and new}, pages 93--111, 2009.

\bibitem[Vince(2002)]{vince2002framework}
Andrew Vince.
\newblock A framework for the greedy algorithm.
\newblock \emph{Discrete Applied Mathematics}, 121\penalty0 (1-3):\penalty0 247--260, 2002.

\bibitem[Xing et~al.(2025)Xing, Xia, Zhang, Chen, Yu, Liu, Liu, Wang, Shan, and Wong]{xing2025dynamicrafter}
Jinbo Xing, Menghan Xia, Yong Zhang, Haoxin Chen, Wangbo Yu, Hanyuan Liu, Gongye Liu, Xintao Wang, Ying Shan, and Tien-Tsin Wong.
\newblock Dynamicrafter: Animating open-domain images with video diffusion priors.
\newblock In \emph{European Conference on Computer Vision}, pages 399--417. Springer, 2025.

\bibitem[Yang et~al.(2023)Yang, Du, Ghasemipour, Tompson, Schuurmans, and Abbeel]{yang2023learning}
Mengjiao Yang, Yilun Du, Kamyar Ghasemipour, Jonathan Tompson, Dale Schuurmans, and Pieter Abbeel.
\newblock Learning interactive real-world simulators.
\newblock \emph{arXiv preprint arXiv:2310.06114}, 1\penalty0 (2):\penalty0 6, 2023.

\bibitem[Zhang et~al.(2022)Zhang, Li, Liu, Zhang, Su, Zhu, Ni, and Shum]{zhang2022dino}
Hao Zhang, Feng Li, Shilong Liu, Lei Zhang, Hang Su, Jun Zhu, Lionel~M Ni, and Heung-Yeung Shum.
\newblock Dino: Detr with improved denoising anchor boxes for end-to-end object detection.
\newblock \emph{arXiv preprint arXiv:2203.03605}, 2022.

\bibitem[Zhang et~al.(2023)Zhang, Wang, Zhang, Zhao, Yuan, Qing, Wang, Zhao, and Zhou]{2023i2vgenxl}
Shiwei Zhang, Jiayu Wang, Yingya Zhang, Kang Zhao, Hangjie Yuan, Zhiwu Qing, Xiang Wang, Deli Zhao, and Jingren Zhou.
\newblock I2vgen-xl: High-quality image-to-video synthesis via cascaded diffusion models, 2023.

\bibitem[Zheng et~al.(2024)Zheng, Peng, Yang, Shen, Li, Liu, Zhou, Li, and You]{opensora}
Zangwei Zheng, Xiangyu Peng, Tianji Yang, Chenhui Shen, Shenggui Li, Hongxin Liu, Yukun Zhou, Tianyi Li, and Yang You.
\newblock Open-sora: Democratizing efficient video production for all, March 2024.
\newblock URL \url{https://github.com/hpcaitech/Open-Sora}.

\bibitem[Zhou et~al.(2024)Zhou, Du, Chen, Li, Yeung, and Gan]{zhou2024robodreamer}
Siyuan Zhou, Yilun Du, Jiaben Chen, Yandong Li, Dit-Yan Yeung, and Chuang Gan.
\newblock Robodreamer: Learning compositional world models for robot imagination.
\newblock \emph{arXiv preprint arXiv:2404.12377}, 2024.

\end{thebibliography}
}

\newpage
\appendix
\section{Appendix}
\label{sec:appendix}


\subsection{Additional Details on World Specification}

We provide implementation details for the pre-processing module. These components ensure consistency across different models and support metric-specific evaluation.

\noindent \textbf{Pre-Processing} We resize all reference images $I$ to a fixed resolution of $640 \times 480$. For the task prompt $L$, we use the aligned step-level caption corresponding to the current sub-action. To ensure viewpoint and temporal consistency, we append the following constraint to the prompt:

\begin{quote}
\textit{“Keep the first-person view of the robot unchanged. Keep the first frame of this video unchanged.”}
\end{quote}

\noindent \textbf{Video Normalization} All generated videos are resized to $640 \times 480$ and resampled to 30 FPS. The normalized video is used to evaluate task scene consistency and semantic alignment.

\noindent \textbf{Trajectory Extraction} We extract 2D trajectories of both end-effectors using a consistent detection pipeline. Specifically, we apply a fine-tuned \textbf{YOLO-World}~\cite{cheng2024yolo} model for per-frame detection and use \textbf{BoT-SORT} ~\cite{aharon2022bot}for temporal association. To ensure fair evaluation, we compute the convex hull of each hand’s trajectory and select the hand with the largest spatial extent—measured by the maximum Euclidean distance between any two points on the hull—as the primary trajectory for motion evaluation.

\noindent \textbf{YOLO-World Training Details.} We fine-tune \texttt{yolov8s-worldv2} on 1451 manually annotated frames from Agibot-World. Two tasks (\textit{Freezer Restocking} and \textit{Factory Packing}) are held out for validation to construct a hard-sample set. The model is trained for 100 epochs and achieves a final performance of \textbf{Recall: 0.91667}, \textbf{Precision: 1.0}.

\subsection{Additional Details on Dataset Curation}

\subsubsection{Task Selection and Scene Diversity}

To support evaluation across a wide range of manipulation scenarios, we select 10 representative tasks from the Agibot-World~\cite{AgiBot2024agibotworld} dataset. The selection prioritizes tasks with clear operational goals, sequential dependencies, and diverse object interactions. These tasks span both household and industrial contexts and are designed to challenge models across spatial reasoning, tool-use, and action ordering.

\begin{itemize}
    \item Retrieving toast from a toaster (\textbf{Toaster})
    \item Pouring water (\textbf{Pour Water})
    \item Setting cutlery (\textbf{Place Cutlery})
    \item Restocking a freezer (\textbf{Restock Freezer})
    \item Producing ice (\textbf{Produce Ice})
    \item Packing laundry detergent (\textbf{Factory Packing})
    \item Cleaning bottles (\textbf{Brush Bottle})
    \item Heating food in a microwave (\textbf{Heat Food})
    \item Installing a showerhead (\textbf{Hang Showerhead})
    \item Storing objects in a drawer (\textbf{Store in Drawer})
\end{itemize}

\begin{figure}[ht]
    \centering
    \includegraphics[width=1\linewidth]{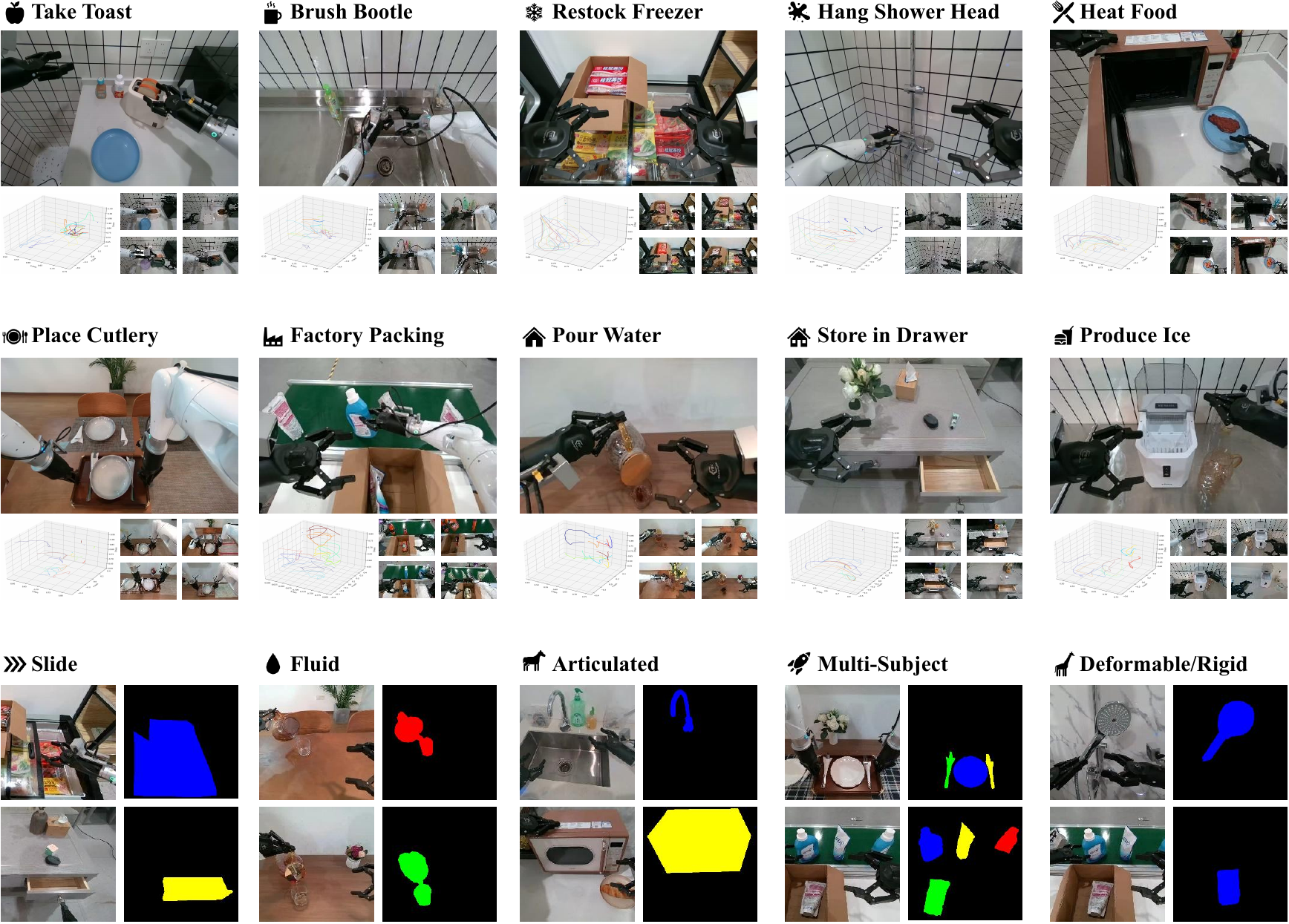}
    \caption{\textbf{Dataset overview.} The first two rows display selected task scenarios and their associated action motion trajectories. The third row categorizes object properties (e.g., fluid, articulated, rigid, deformable, multi-body) using color-coded legends and representative examples.}
    \label{fig:dataset dashboard}
\end{figure}

These tasks exhibit substantial variation in manipulated object types (e.g., rigid, deformable, articulated), spatial layouts, and interaction complexity. Figure~\ref{fig:dataset dashboard} visualizes the task scenes, their corresponding trajectory overlays, and object property distributions.

\subsubsection{Action Motion Sampling Strategy}

To construct a comprehensive and diverse evaluation set, we uniformly sample 100 video instances per task. Each task is decomposed into fine-grained primitive actions, ensuring one-to-one correspondence between video segments, sub-action labels, and textual descriptions.

For trajectory analysis, we extract the left and right end-effector positions and convert them into voxel grid representations. A pairwise similarity matrix is computed using 3D Intersection over Union (IoU):

\begin{equation}
\text{IoU}_{i,j} = \frac{
    \sum \min(V^L_i, V^L_j) + \sum \min(V^R_i, V^R_j)
}{
    \sum \max(V^L_i, V^L_j) + \sum \max(V^R_i, V^R_j) + \epsilon
}
\end{equation}

where \(V^L\) and \(V^R\) denote the voxel grids for the left and right end-effectors, respectively.

To promote trajectory diversity in the final evaluation subset, we adopt a greedy selection algorithm~\cite{vince2002framework}:

\begin{itemize}
    \item Start by selecting the trajectory with the lowest average IoU relative to all others.
    \item Iteratively select the trajectory with the lowest average IoU to the already selected set.
    \item Repeat until 10 representative trajectories are selected for each task.
\end{itemize}

This sampling strategy ensures that the evaluation set includes both common and atypical motion patterns, providing a broad spectrum of behavioral diversity to test the generalization ability of generative models. The resulting distribution is reflected in Figure.~\ref{fig:dataset dashboard}.



\subsection{Additional Details on Metrics}



\subsubsection{Scene Consistency: DINOv2 vs. VBench Analysis}

\noindent \textbf{Fine-tuning DINOv2 for Embodied Scenes.}
To enhance the extraction of task-relevant visual features in embodied operation scenarios, we fine-tune DINOv2~\cite{oquab2023dinov2} on the Agibot-World dataset using 20,000 iterations of unsupervised training. We use the \texttt{dinov2-vitb14-reg4} checkpoint as the initialization. We visualize feature maps from three model variants: (1) the original DINO~\cite{zhang2022dino} (ViT-B/16, used in VBench), (2) pre-trained DINOv2, and (3) our fine-tuned DINOv2.

\begin{figure}[t]
    \centering
    \includegraphics[width=1.0\linewidth]{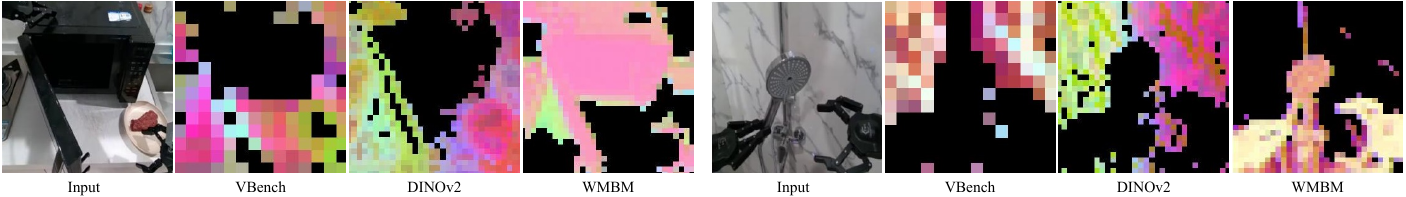}
    
    \caption{\textbf{Feature map comparison across models.} DINOv2 fine-tuned on embodied data captures agents and tools with sharper spatial coherence, enabling more reliable scene stability evaluation.}
    \label{fig:feature_map_comparison}
\end{figure}

As shown in Figure.~\ref{fig:feature_map_comparison}, only the fine-tuned DINOv2 consistently focuses on task-relevant agents and manipulated tools, while the others fail to highlight key foreground regions or exhibit background-foreground entanglement. This justifies the necessity of adapting foundation models to the embodied task domain.

\begin{figure}[t]
    \centering
    \includegraphics[width=1.0\linewidth]{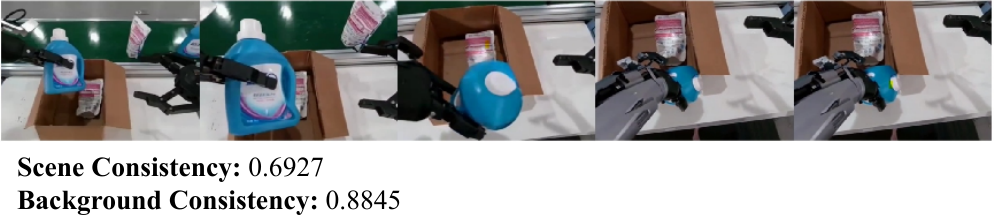}
    
    \caption{\textbf{Scene Consistency failure cases.} Despite camera movement and background drift, VBench assigns high scores (0.88–0.91). Our metric, however, detects the instability through decreased cosine similarity.}
    \label{fig:scene_consistency_cases}
\end{figure}

\noindent \textbf{Failure Cases of VBench Background Consistency.}
We further visualize representative cases in which VBench assigns high background consistency scores despite significant viewpoint changes or layout shifts. As shown in Figure.~\ref{fig:scene_consistency_cases}, the cosine similarity of our metric drops significantly in these cases, whereas VBench remains insensitive. This highlights the limitation of VBench's DINO-ViT-B/16 features in separating foreground and background cues.

\subsubsection{Action Motion Metrics}

\noindent \textbf{Symmetric Hausdorff Distance Consistency(HSD Consistency)}
    
The Symmetric Hausdorff Distance\cite{serra1998hausdorff} (SymH) measures the maximum spatial deviation between the generated trajectory and the corresponding ground truth trajectory. This distance represents the greatest of the minimum distances between points from both trajectories. HSD is particularly useful for evaluating the spatial alignment of the generated trajectory with the true trajectory, ensuring that the generated path adheres to expected movement patterns and does not deviate significantly from the true action trajectory.

To ensure that the score is positively correlated with consistency, we take the reciprocal of this value:

\begin{equation}
\text{HSD}_{\text{score}} = \frac{1}{d_{\text{symH}}(G, P)}
\end{equation}
\text{Where } G \text{ represents the ground truth trajectory, and } P \text{ represents the generated trajectory.}
    
\noindent \textbf{Normalized Dynamic Time Warping Distance Consistency(NDTW Consistency)}

NDTW\cite{muller2007dynamic} is used to evaluate the overall shape similarity and temporal alignment of trajectories. While HSD focuses on spatial deviations, NDTW evaluates the overall trajectory shape and how well the generated actions align with the timing and sequence of the true actions. This metric is particularly useful for capturing the temporal causality and task sequencing that the model should learn from the true trajectory.

By aligning both the spatial and temporal dimensions of the trajectories, NDTW assesses whether the generated sequence matches the timing and order of the ground truth actions. The similarity score is then calculated as the reciprocal of this value:

\begin{equation}
\text{NDTW}_{\text{score}} = \frac{1}{\text{NDTW}(G, P)}
\end{equation}

\noindent \textbf{Dynamic Consistency (DYN)}
    
Velocity and acceleration are critical components in robotic control, directly affecting the physical feasibility of the generated actions. To evaluate how well the predicted trajectories align with the real-world motion characteristics, we extract the 2D velocity and acceleration time series from both predicted and ground-truth trajectories. We then compute the Wasserstein distance\cite{villani2009wasserstein} (Earth Mover’s Distance, EMD) between these distributions to quantify their differences.

 The Wasserstein distance captures the global distributional alignment between sequences, it allows for soft matching and does not require strict temporal alignment, which makes it more robust and better suited to capturing continuous motion trends across the entire trajectory.

To enhance the robustness of this metric across different motion amplitudes, inspired by the IoU calculation method, we introduce an amplitude normalization factor that uses the difference between the maximum and minimum velocity/acceleration to construct the following ratio:
    \begin{equation}
    \text{VR} = \frac{
        \min\left( \max(v^{\text{gt}}) - \min(v^{\text{gt}}),\ \max(v^{\text{pred}}) - \min(v^{\text{pred}}) \right) + \epsilon
    }{
        \max\left( \max(v^{\text{gt}}) - \min(v^{\text{gt}}),\ \max(v^{\text{pred}}) - \min(v^{\text{pred}}) \right) + \epsilon
    }
    \end{equation}
    \begin{equation}
    \text{AR} = \frac{
        \min\left( \max(a^{\text{gt}}) - \min(a^{\text{gt}}),\ \max(a^{\text{pred}}) - \min(a^{\text{pred}}) \right) + \epsilon
    }{
        \max\left( \max(a^{\text{gt}}) - \min(a^{\text{gt}}),\ \max(a^{\text{pred}}) - \min(a^{\text{pred}}) \right) + \epsilon
    }
    \end{equation}
    where $ \epsilon = 1 \times 10^{-8} $.

To account for variations in motion amplitudes, we introduce amplitude normalization factors. The final dynamic consistency score is defined as:

\begin{equation}
\text{DYN}_{\text{score}} = \alpha \cdot \text{VR} \cdot \frac{1}{W(v)} + \beta \cdot \text{AR} \cdot \frac{1}{W(a)}, \quad 
\alpha = 0.007, \quad \beta = 0.003
\end{equation}

Here, $W(\cdot)$ represents the Wasserstein distance, and $ \text{VR}$, $\text{AR}$ are the amplitude normalization factors for velocity and acceleration, respectively. These corrections ensure that low-amplitude trajectories do not introduce numerical amplification, thereby preserving the accuracy of dynamic consistency assessment.

\subsection{Visual Examples of Model Generation Results}

To complement the discussion in Section~\ref{sec:discussion}, we provide qualitative examples illustrating the observed characteristics of recent SOTA models evaluated under WMBM.

\begin{figure}[h]
  \centering
  \includegraphics[width=1\linewidth]{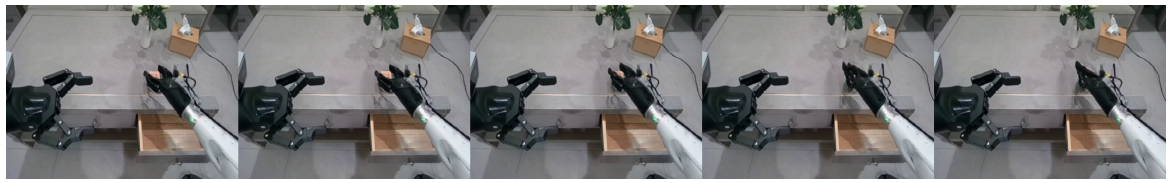}
  
  \caption{\textbf{Domain-adapted model failure case.} The robot hand moves toward the correct region but fails to close the gripper on the object, resulting in empty grasping.}
  \label{fig:adapted_emptygrasp}
\end{figure}

\noindent \textbf{Domain-Adapted Models.}
Despite strong semantic alignment, domain-adapted models sometimes fail in action grounding. As shown in Figure.\ref{fig:adapted_emptygrasp}, the generated video depicts a precise scene and goal instruction, but the agent executes an empty grasp motion without interacting with the object.

%

%

    



\begin{figure}
    \centering
    \includegraphics[width=1\linewidth]{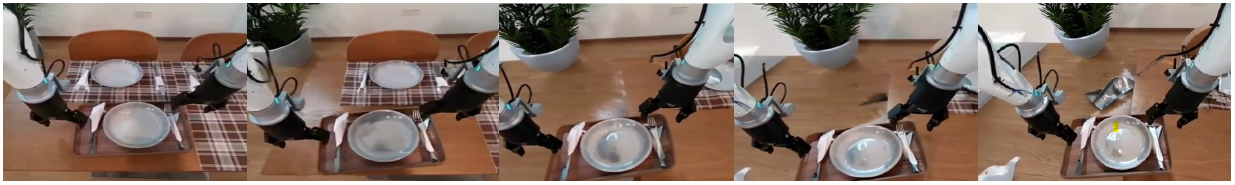}
    \caption{Despite being given explicit camera viewpoint control instructions, COSMOS fails to maintain a consistent viewpoint throughout the video. This indicates a limitation in its ability to follow spatial constraints, leading to unstable or drifting perspectives.}
    \label{fig:cosmos_camera_control}
\end{figure}

\begin{figure}
    \centering
    \includegraphics[width=1\linewidth]{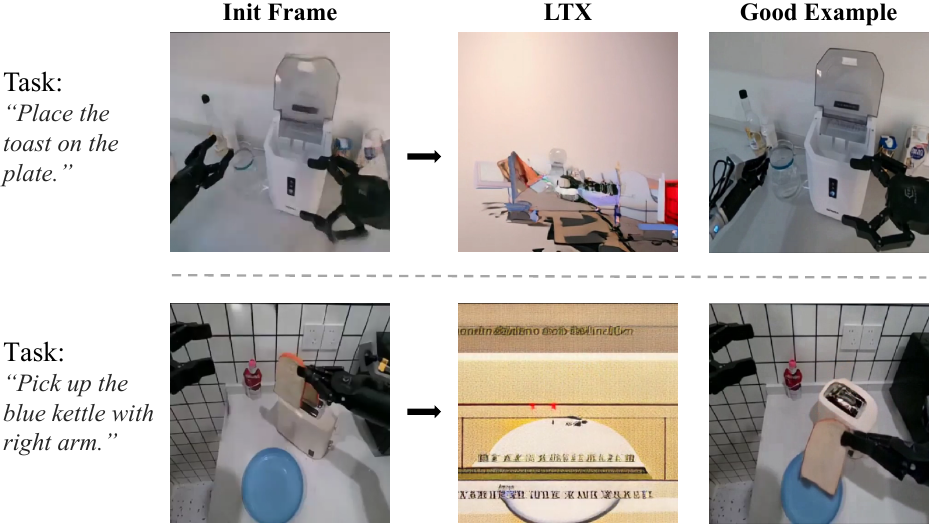}
    
    \caption{Examples illustrating the poor task understanding and temporal instability of the LTX model. The middle column shows LTX-generated frames, which often exhibit abrupt visual changes and scene inconsistencies. In contrast, the rightmost column presents examples with better scene preservation, highlighting the gap in temporal coherence.}
    \label{fig:ltx_failures}
\end{figure}

\begin{figure}
    \centering
    \includegraphics[width=1\linewidth]{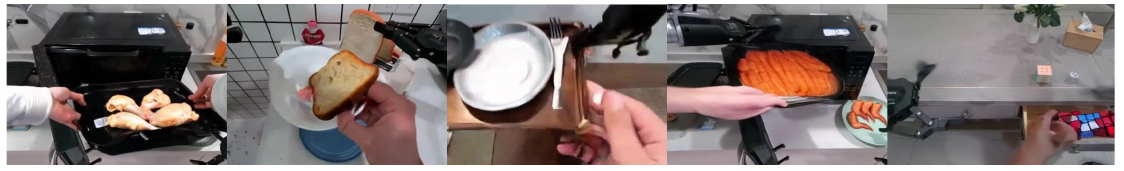}
    
    \caption{Generated videos from COSMOS and LTX models often depict human hands instead of robotic arms, indicating a bias toward human hand representations in their training data. This bias hinders the models' ability to correctly generalize to robotic manipulation tasks, where accurate mapping to robotic arms is essential.}
    \label{fig:cosmos_humanhand}
\end{figure}

\noindent \textbf{LTX and COSMOS.}
Figure.~\ref{fig:ltx_failures} shows LTX generating abrupt scene transitions and failing to maintain object continuity. LTX and COSMOS, as in 
Figure.~\ref{fig:cosmos_humanhand}, frequently renders human hands instead of robot arms, revealing a semantic adaptation failure.A failure of viewpoint control in COSMOS is illustrated in Figure.~\ref{fig:cosmos_camera_control}.

\begin{figure}[thp]
    \centering
    \includegraphics[width=1\linewidth]{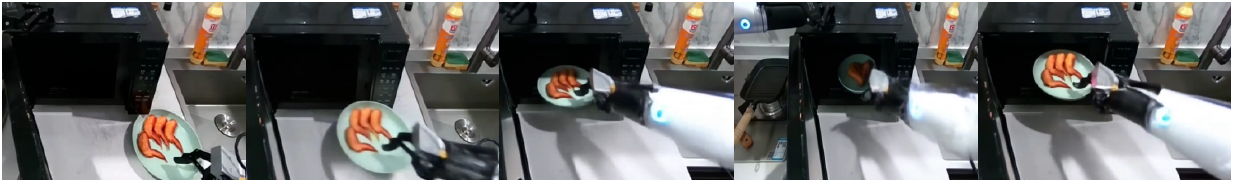}
    
    \caption{An example from OpenSora illustrating unstable robotic arm motion. Although the scene is semantically aligned with the manipulation task, the arm exhibits visible jitter and lacks smooth trajectory control, consistent with the motion instability discussed in the main text.}
    \label{fig:opensora}
\end{figure}

\noindent \textbf{OpenSora.}As discussed in the main text, OpenSora shows partial understanding of task semantics but struggles with motion control. Figure.~\ref{fig:opensora} presents a representative example highlighting this issue: while the generated scene correctly reflects the intended manipulation context, the robotic arm undergoes significant jitter and fails to execute smooth movements. This supports our observation that OpenSora's motion instability remains a key limitation in embodied video generation.

\end{document}